\documentclass[letterpaper, 10 pt, conference]{ieeeconf}  

\IEEEoverridecommandlockouts                              

\usepackage{graphics} 
\usepackage{subfigure}
\usepackage{polynom}
\usepackage{fixltx2e}
\usepackage{hyperref}
\usepackage{textgreek}
\usepackage{caption}
\usepackage{float}
\usepackage{siunitx}
\usepackage{epsfig} 
\usepackage[ruled]{algorithm2e} 
\usepackage{url}
\usepackage{multirow}
\usepackage{dblfloatfix}
\usepackage{amsmath} 
\usepackage{amssymb}  

\usepackage[space]{cite}

\title{\LARGE \bf
Affordance-Based Mobile Robot Navigation Among Movable Obstacles
}

\author{Maozhen Wang$^{1}$, Rui Luo$^{1}$, Aykut {\"O}zg{\"u}n {\"O}nol$^{1}$ and Ta\c{s}k{\i}n~Pad{\i}r$^{1}$
\thanks{This research is supported by the National Science Foundation under Award Nos. 1544895, 1944453, 1928654.}
\thanks{$^{1}$Institute for Experiential Robotics, Northeastern University, Boston, MA 02115, USA
       {\tt\small \{wang.mao, luo.rui, onol.a, t.padir\}@northeastern.edu}}}

\begin{document}

\setlength{\textfloatsep}{6pt}
\captionsetup[table]{name=TABLE,justification=centering,labelsep=newline,textfont=sc}

\maketitle
\thispagestyle{empty}
\pagestyle{empty}

\begin{abstract}
Avoiding obstacles in the perceived world has been the classical approach to autonomous mobile robot navigation. However, this usually leads to unnatural and inefficient motions that significantly differ from the way humans move in tight and dynamic spaces, as we do not refrain interacting with the environment around us when necessary. Inspired by this observation, we propose a framework for autonomous robot navigation among movable obstacles (NAMO) that is based on the theory of affordances and contact-implicit motion planning. We consider a realistic scenario in which a mobile service robot negotiates unknown obstacles in the environment while navigating to a goal state. An affordance extraction procedure is performed for novel obstacles to detect their movability, and a contact-implicit trajectory optimization method is used to enable the robot to interact with movable obstacles to improve the task performance or to complete an otherwise infeasible task. We demonstrate the performance of the proposed framework by hardware experiments with Toyota's Human Support Robot.
\end{abstract}

\section{Introduction}
Personal service robots are on the verge of being ubiquitous as a key technology to provide independent living at home for senior citizens and individuals with disabilities. Tasks such as fetching an object, tidying up a room, storing groceries, and social interactions are among the desired capabilities of personal service robots. For a mobile robot to perform such physical manipulation tasks in a human environment, reliable autonomous navigation in dynamic and cluttered environments is a fundamental capability and remains to be a grand challenge. As home environments are rarely static due to moving items, such as bags, laundry baskets, and stools, the traditional navigation paradigm relying on passive obstacle avoidance result in ineffective and infeasible outcomes. These include longer detours and failure to find a navigable path in cluttered spaces. On the other hand, humans are capable of handling unknown obstacles in an interactive way by pushing around or removing them to create free spaces to navigate. Hence, such physical interactions with the environment are necessary for mobile robots to perform complex service tasks successfully as cohabitants in the home.

When a human wants to move an unknown obstacle to clear the path, two consecutive steps are carried out: (\textit{i}) the \textit{movability} of the object is assessed by trying to move it and (\textit{ii}) depending on the object’s movability, it is either taken out of the way or treated as a static obstacle. While there is an abundance of previous research and off-the-shelf tools for navigation by avoiding obstacles, there are only a few studies for navigation by interacting with obstacles, which is called navigation among movable obstacles (NAMO) in the related literature \cite{5649744, 7041342, biye}. Nevertheless, the exploration of objects by the robot to make informed decisions about interacting with them is still an open research problem.

   \begin{figure}[tpb]
      \centering
      \includegraphics[width=0.476\textwidth]{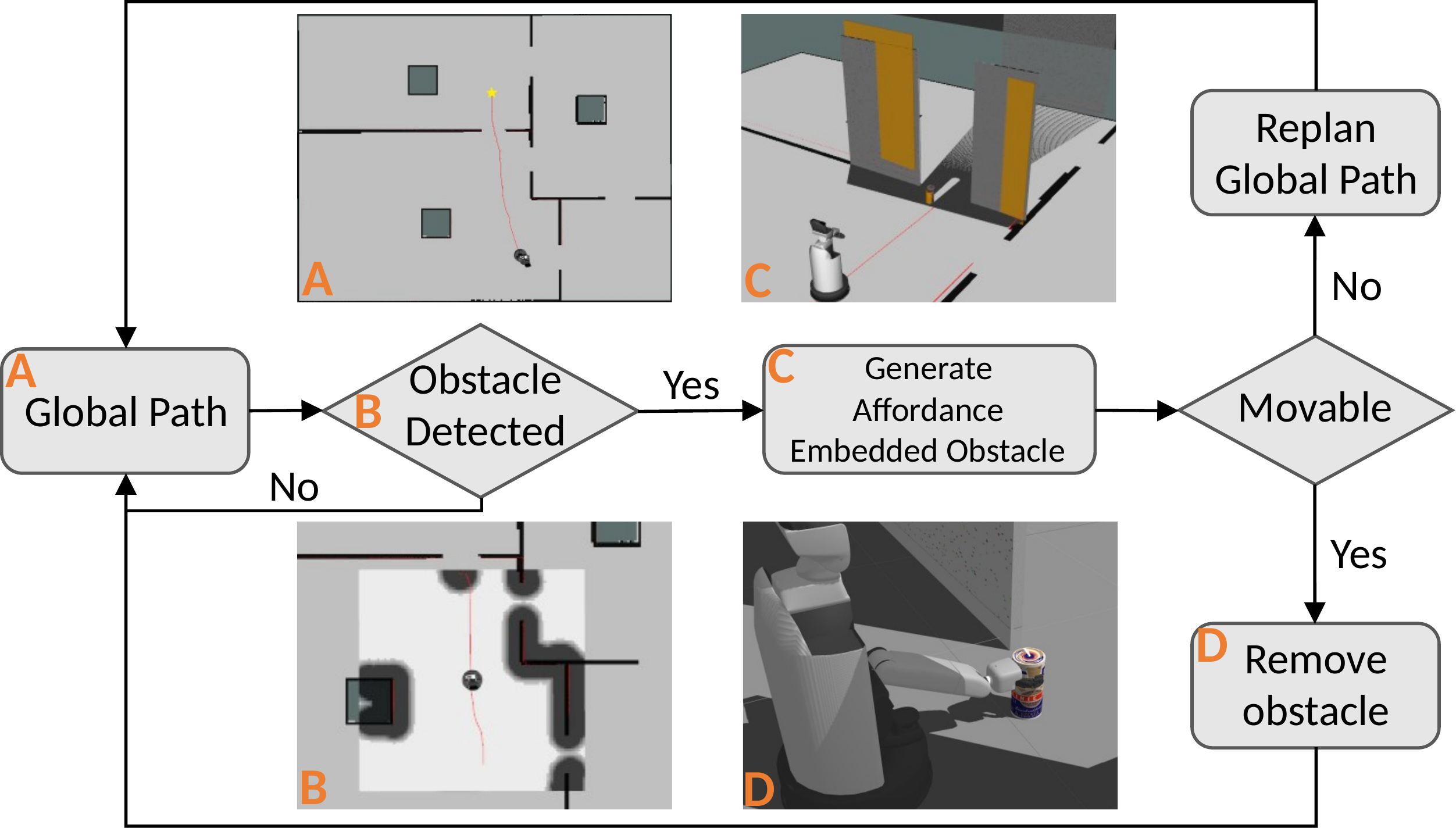}
      \caption{An overview of the proposed affordance-based navigation algorithm. Given a global map and a goal state to navigate, first, a collision-free path is planned (A). While executing this path, the robot keeps checking for unregistered obstacles blocking its way (B). If such an obstacle is detected, the robot performs an affordance assessment and validation for the obstacle to determine its \textit{movability} (C). If the object is \textit{movable}, the robot removes it from its path and pursues the initially planned path (D). Otherwise, the robot plans a new global path that avoids the \textit{unmovable} obstacle.}
      \label{overview}
   \end{figure}

Existing work on NAMO usually take the \textit{movability} as pre-knowledge or infer it through an object recognition method. However, humans can determine the \textit{movability} of an object without precise knowledge of its physical properties by extracting an affordance measure for the object. The concept of affordances was first introduced by the psychologist J.J. Gibson\cite{affordance}. In essence, the affordance describes the actions that an object can afford to the perceiving agent. In the case of navigation, affordance can be simplified to \textit{movability}, or more specifically to \textit{pushability} or \textit{liftability} of the obstacle. For example, a chair affords being pushed away and a cup affords being lifted. Inspired by this, we propose a navigation algorithm for a mobile manipulator to explore and utilize the \textit{movability} of unexpected obstacles through an affordance embedding method. An overview of the navigation pipeline is presented in Fig. \ref{overview}. Affordances are used such that the robot can handle novel objects in a more flexible way and thus can persistently navigate in dynamic environments.


\subsection{Related Work}
Affordance offers robots to perceive the environment in a way similar to humans do, by capturing the properties of objects without identifying them. Min et al. \cite{survey} provide an overview of affordance research in developmental robotics and show that affordance is useful in planning, control, and perception. In \cite{doi:10.1002/rob.21546}, pre-defined affordance templates are used to fit objects to be manipulated. This approach has been proven to be effective for manipulation tasks during DARPA Robotics Challenge \cite{7759708}.

Recent research, on the other hand, focuses on perceiving affordance for unknown environments rather than the template-based approach which only works for pre-defined objects. Myers et al. \cite{7139369} show that geometry is critical for predicting the affordance for novel objects. In \cite{DBLP:journals/corr/abs-1709-07326}, a convolutional neural network is used to detect affordance from RGB images. Instead, \cite{P1} proposes using point cloud data for this purpose. Affordances are inferred from primitive geometry shapes that compose the object. This method is more generalizable since it does not require any training. \cite{P2} extends this work by combining basic affordances to compose complex affordances, such as bimanual affordances. The affordance detection method adapted from \cite{P1} is further developed and applied to solve NAMO problems by grouping independent affordance and object information to form what we call affordance-embedded obstacle in this study.

While affordance-based methods have been extensively studied for robotic manipulation, the literature is relatively limited for navigation. {\c{S}}ahin et al. \cite{doi:10.1177/1059712307084689} formalize affordance expressions for robot control. Using this formalization, \cite{doi:10.1177/1059712310370625} perceives the traversability from the environment and navigate the robot based on traversable directions. \cite{7081123} uses affordance for navigation by building topological maps based on the predicted affordances of objects. However, none of these studies considers moving the objects but just avoiding them.

Motion planning among movable obstacles is a difficult problem, as discussed in \cite{Wilfong1991}. An algorithm to efficiently plan a path in an unknown environment consisting of movable rectangular obstacles is presented in \cite{5649744} and is further optimized and extended to arbitrary shaped obstacles in \cite{7041342}. While this algorithm gives promising results in a 2D cell-decomposed and axis-aligned environment, the detection of \textit{movability} from sensory data is not studied. Similarly, in \cite{biye}, a sampling-based planner is proposed for NAMO but it assumes pre-knowledge of the \textit{movability} of objects. \cite{8452226} proposes a more complete pipeline that consists of movable object detection and clearing. This is closely aligned with the objective of our study but in \cite{8452226} \textit{movability} of objects is inferred through object recognition, which limits the applicability of the framework, especially for novel objects, and lacks the mass properties of the object, as only vision is used. In this paper, we take advantage of the affordance phenomenon and propose a generalized NAMO framework that can handle realistic domestic scenarios.

\subsection{Contributions}
The main contributions of our work are:
\begin{itemize}
    \item a modeling method for embedding unknown objects with affordance information that is obtained through physical interactions with them;
    \item an efficient decision-making algorithm that enables the robot to deal with unexpected and unknown obstacles by exploiting their affordances; and
    \item a contact-implicit motion planner customized for navigation such that a contact-interaction trajectory can be planned given only a goal pose for the robot.
\end{itemize}
These salient features are incorporated in a navigation pipeline with collision-free path planning, object detection, object pick-and-place, and path resumption components.

\section{Methodology}
\subsection{Problem Specification}
This research is aimed at solving the robot navigation problem in a known map updated with randomly placed unknown objects. Typical use cases include navigation in the home, or common environments, such as hospitals, libraries, museums, or elder care facilities. Given a global map and a goal position, a feasible path reaching the goal can be planned. However, this path may be blocked by various obstacles, for instance, by a chair left out or a shipping package just received. Without knowing how to interact with such unknown items, the robot has to plan a detour or may be immobilized if the only path is now blocked. Since the original path took all known static obstacles into account, a straightforward solution is to remove the blockage. Although it is uncertain if removing the blockage is always better than avoiding, our approach enables the robot to reason about physical interaction with the environment. This ability is crucial especially in immobilized situations. To achieve that, the proposed framework performs three main steps: (\textit{i}) determining the object that needs to be moved, (\textit{ii}) assessing the \textit{movability} of the object, and (\textit{iii}) planning and executing a motion that either removes the object or avoids it.

Two types of \textit{movability} are considered for obstacles: \textit{liftability} and \textit{pushability}. A \textit{liftable} object can be removed by picking it up using the robot's gripper. To take advantage of the latter affordance, we would like to enable the robot to navigate to the goal by making and breaking contacts with the environment, rather than moving an obstacle to a specified configuration. For this purpose, we use a contact-implicit trajectory optimization (CITO) method that is built upon our previous work \cite{onol2018comparative,onol2019contact,onol2020tuning}.

\subsection{Obstacle Embedded with Affordance}
The affordance in our NAMO framework is referred as the \textit{movability} of the obstacle, and its extraction is based on the method presented in \cite{P1}. We recall the method here and introduce our proposed modifications to it. The affordance extraction is achieved by two steps: visual detection and validation. The input to the system is a registered point cloud. A part-based segmentation \cite{stein2014convexity} is then performed to divide points into different clusters. Each cluster can be regarded as one obstacle. For each point-cloud cluster, geometric primitives (i.e., cylinder, plane, or sphere) of various size can be extracted using a sample consensus method such as RANSAC \cite{10.1145/358669.358692}. The procedure of primitive extraction will be applied iteratively to the point-cloud cluster so that an irregular shaped obstacle can be disassembled as a group of primitives. This detection approach thus has a good generalization ability without the need for training.

The extracted primitives are assigned with hypotheses of the way how the robot can interact with the obstacle and thus can be regarded as affordance hosts. For each host, the robot needs to perform respective actions to validate the assigned hypothesis. Our rules to assign and validate affordance hypotheses are detailed in Table \ref{table1}. We considered only \textit{pushability} and \textit{liftability} in this study as lifting and pushing are among the most common actions humans take to clear a path. Referring to the last column in Table \ref{table1}, the two validation methods are:
\begin{enumerate}
  \item Grasp the primitive and apply vertical force to lift. Compare the resistance force against the maximum value $f_L$.
  \item Try pushing the primitive using the robot's gripper. Compare the resistance force against the maximum value $f_P$.
\end{enumerate}
The values $f_L$ and $f_P$ are robot-specific, chosen considering the maximum effort limits of the actuators.

\begin{table}[t]
\renewcommand{\arraystretch}{1.3}
\vspace{0.2cm}
\centering
\caption{Rules for hypothesizing and validating affordance}
\label{table1}
{\begin{tabular}{l}
\begin{tabular}{|l|l|l|c|c|}
\hline
\textbf{Primitive}  & \textbf{Affordance}   & \textbf{Parameters}    & $\textbf{Conditions}^1$    & \textbf{Valid.} \\ \hline
\multirow{2}{*}{\begin{tabular}[c]{@{}c@{}} Cylinder \end{tabular}} 
                & Lift & \begin{tabular}[c]{@{}c@{}}Radius $r$ \end{tabular} & $r < c_1$ & 1 \\ 
    \cline{2-5} & Push & \begin{tabular}[c]{@{}l@{}}Radius $r$ \end{tabular} & $r < c_2$ & 2 \\ \hline
\multirow{2}{*}{\begin{tabular}[c]{@{}c@{}} \vspace{2mm} \\ Plane \end{tabular}}
                & Lift & \begin{tabular}[c]{@{}l@{}}Width $d$\\ Height $h$\\ Normal $\mathbf{n}$ \end{tabular} & \begin{tabular}[c]{@{}c@{}} $d < c_3$ \\ $\mathbf{n} \perp \mathbf{z}_W$\end{tabular} & 1  \\ 
\cline{2-5}     & Push & \begin{tabular}[c]{@{}l@{}}Width $d$\\ Height $h$\\ Normal $\mathbf{n}$ \end{tabular} & \begin{tabular}[c]{@{}c@{}} $d \times h < c_4$ \\ $\mathbf{n} \perp \mathbf{z}_W$\end{tabular} & 2 \\ \hline
\multirow{2}{*}{Sphere}
                & Lift & Radius $r$ & $r < c_5$ & 1 \\ 
\cline{2-5}     & Push & Radius $r$ & $r < c_6$ & 2 \\ \hline
\end{tabular}\\
\rule{0in}{1.2em}
$^1$\scriptsize The values $c_i$ are robot-specific constants. $\mathbf{z}_W$ is the $\mathbf{z}$ axis of the world frame.
\end{tabular}}
\end{table}

To better serve the navigation purposes, we make two modifications to the original method in \cite{P1}. First, instead of considering each geometric primitive individually, we group those that belong to the same obstacle, so that detected affordances are assigned to the obstacle that needs to be moved to clear the path. Once this assignment is done, the obstacle is called an affordance-embedded obstacle. This affordance-obstacle binding is crucial for navigation as we are interested in modifying the configuration of the obstacles if necessary to reach the goal pose, rather than using individual affordances to complete the task. In practice, such bindings make the geometry of the obstacle available to the robot after the affordance validation. The geometry of the obstacle is necessary for both marking \textit{unmovable} obstacle on the map and running CITO to find a pushing motion. Specifically, the footprint of the obstacle is important as the robot uses its base to contact the obstacle.
Moreover, the cost of the validation process can be reduced substantially by grouping the primitives since all affordance hypotheses of the same obstacle share the same validation result. For instance, if one vertical surface of an obstacle is pushable, then its all other vertical surfaces should also be pushable assuming no constraints due to other obstacles. Nevertheless, if there are such constraints, they will be taken into account in CITO. Once the validation phase is complete, the robot can predict the outcome of interacting with different obstacles and take only necessary actions during navigation.

The second modification is that we rank the affordance hypotheses by the ease of utilization during navigation. If geometrically admissible, the robot would first validate \textit{liftability} since picking up the obstacle immediately clears the path. \textit{Pushability} would be validated only if the obstacle is already determined to be \textit{unliftable}. By ranking the hypotheses, the affordance validation process is expedited as fewer attempts are required.

An affordance-embedded obstacle is formalized as $\mathbb{O}$\{footprint, $\psi_i$(shape, parameters, affordance)\}. The footprint of an obstacle is calculated by projecting its point cloud onto the ground. For collision-checking purposes, we consider the convex hull of each obstacle. The $i^{th}$ geometric primitive of the affordance-embedded obstacle, $\psi_i$ is defined by three attributes, i.e., shape, parameters, and affordance. Parameters include the dimensional parameters of the corresponding shape as listed in Table \ref{table1} as well as the center position of the primitive. Affordance describes whether the primitive is \textit{liftable}, \textit{pushable}, or \textit{unmovable}.

\subsection{Contact-Implicit Trajectory Optimization}
To exploit the \textit{pushability} affordance, a customized version of the CITO method presented in \cite{onol2019contact} is used. The difference of our version is that the goal is specified in terms of the robot's pose, rather than the object's pose. The robot is allowed to make/break contacts with \textit{movable} obstacles using its holonomic cylindrical base as necessary to reach the goal. In this approach, a variable smooth contact model (VSCM) \cite{onol2018comparative} is employed to enable gradient-based optimization to reason about contacts. The VSCM is based on exerting smooth virtual forces upon the free bodies in the environment (i.e., \textit{movable} objects) and letting the optimization to adjust the amount of virtual forces such that they vanish once a motion that completes the task using physical forces is found.

In this case, the state vector $\mathbf{x}\in\mathbb{R}^{n}$ consists of the configurations and velocities of the robot's base and $n_o$ \textit{movable} obstacles, i.e., $\mathbf{x} \triangleq [\mathbf{q}_r^T,\mathbf{q}_o^T,\dot{\mathbf{q}}_r^T,\dot{\mathbf{q}}_o^T]^T$ where $\mathbf{q}_r\in\mathbb{R}^3$ and $\mathbf{q}_o\in\mathbb{R}^{3n_o}$ are the configuration vectors for the robot and the \textit{movable} objects in $SE(2)$. We define $n_p$ contact pairs between the mobile base and the vertical surfaces of \textit{movable} obstacles. For each contact pair, there is a virtual force that relates the configuration of the system to the dynamics of the movable objects.

For the $i^{th}$ contact pair, the magnitude of the normal virtual force $\gamma_i$ is calculated by $\boldsymbol{\gamma}_i=k_i e^{-\alpha \boldsymbol{\phi}_i(\mathbf{x})}$, where $k_i$ is the virtual stiffness, $\alpha$ is the curvature, and $\boldsymbol{\phi}_i(\mathbf{x})$ is the signed distance. The generalized virtual force acting upon the center of mass of the movable object associated with the contact pair is then calculated by $\boldsymbol{\lambda}_i(\mathbf{x})=\boldsymbol{\gamma}_i[\mathbb{I}_{3} \ \widehat{\mathbf{l}}_i(\mathbf{x})^{T}]^{T}\mathbf{n}_i(\mathbf{x})$, where $\mathbb{I}_{3}$ is the $3\times3$ identity matrix, $\widehat{\mathbf{l}}_i$ is the skew-symmetric matrix form of the vector from center of mass of the obstacle to the mobile base, and $\mathbf{n}_i$ is the contact surface normal for the contact candidate on the movable object. The net virtual force exerted on a movable object is the sum of $\boldsymbol{\lambda}$ vectors associated with the object. To let the optimization adjust the virtual forces, the virtual stiffness values $\mathbf{k}\in\mathbb{R}^{n_p}$ are defined as a part of the control input $\mathbf{u}$ in addition to the velocities of the holonomic base, i.e., $\mathbf{u} \triangleq [\dot{\mathbf{q}}_r^{T}, \mathbf{k}^{T}]\in\mathbb{R}^{3+n_{p}}$.

The trajectory optimization problem is given by:
\begin{subequations}
    \begin{align}
        \underset{\mathbf{u}_{1,...,N},\mathbf{x}_{1,...,N+1}}{\text{minimize }} C_F(\mathbf{x}_{N+1}) + \sum_{i=1}^{N} C_I(\mathbf{x}_i,\mathbf{u}_i)
    \end{align}
subject to:
    \begin{align}
        & \mathbf{x}_{i+1} = f(\mathbf{x}_i,\mathbf{u}_i) \text{ for } i=1,...,N, \label{eq:dynamics_constraint} \\
        & \mathbf{u}_{L} \leq \mathbf{u}_{1,...,N} \leq \mathbf{u}_{U}, \ \mathbf{x}_{L} \leq \mathbf{x}_{1,...,N+1} \leq \mathbf{x}_{U}, \label{eq:box_constraints}
    \end{align}
\end{subequations}
where $f(\mathbf{x}_i,\mathbf{u}_i)$ describes the nonlinear underactuated dynamics of the system, $N$ is the number of time steps, $C_F$ and $C_I$ are the final and integrated cost terms, and $\mathbf{u}_L$, $\mathbf{u}_U$, $\mathbf{x}_L$, and $\mathbf{x}_U$ are the lower and upper control and state limits.

Since the aim of this optimization is to plan a trajectory that takes the robot to a goal configuration using the shortest path, the final and integrated cost terms, $C_F$ and $C_I$ are defined as follows:
\begin{subequations}
    \begin{align}
        & C_F = w_1 ||{\mathbf{q}}_{r}^{e}||_{2}^{2}, \\
        & C_I = w_2||{\dot{\mathbf{q}}}_{i}||_{2}^{2} + w_3||{\mathbf{k}}_{i}||_{1},
    \end{align}
\end{subequations}
where ${\mathbf{q}}_{r}^{e}$ is the deviation from the navigation goal, ${\dot{\mathbf{q}}}_{i}$ is the vector of velocities of the robot and the movable objects, and $w_1$, $w_2$, and $w_3$ are the weights. Here, the final cost term drives the robot to the goal. The first term in the integrated cost is used to minimize the effort, and the second term penalizes the virtual forces. In this study, the weights are selected as $w_1=2.5 \times 10^{3}$, $w_2=10^{-4}$, and $w_3=7$. The orientation is not considered since a desired orientation can be easily reached by stand turning. In addition to these, the robot needs to avoid collisions with static obstacles. For this purpose, box constrains on states are employed in \eqref{eq:box_constraints}. 

This trajectory optimization problem is then solved using a variant of the successive convexification (SCVX) algorithm originally proposed in \cite{acikmese2016scvx}. Three main steps are repeated in SCVX: (\textit{i}) linearizing the nonlinear dynamics about the previous solution; (\textit{ii}) solving the resulting convex subproblem within a trust region; and (\textit{iii}) updating the trust region radius based on the similarity between linearized and original cost values. In this study, SCVX is slightly modified as proposed in \cite{onol2019contact}. This helps to eliminate the potential accumulation of defects in the state trajectory and allows larger trust regions. Although this modified version does not inherit the convergence guarantees of the original method, we observe faster speed and more reliable convergence in our application.

\subsection{Navigation Framework}

\begin{algorithm}[tb]
\SetAlgoLined
 ${path} \gets A^*(M_{global}, R_{init}, R_{goal})$\;
 \While{$R \neq R_{goal}$}{
  $R \gets TakeStep(path)$\;
  $M_{local} \gets \mathrm{GetLocalMap}(R)$\;
  $o \gets \mathrm{CheckCollision}(M_{local}, path)$\;
  \uIf{$o \neq \varnothing$}{
   $P \gets \mathrm{GetPointCloud}()$\;
   $\Psi \gets \mathrm{PrimitiveDetect}(P)$\;
   $\psi \gets \mathrm{FilterPrimitive}(\Psi, o)$\;
   \uIf{$\mathrm{ValidateLift}(\psi) == True$}{
   $\mathrm{ClearObstacle}(\psi, lift)$\;
   \textbf{Continue}
   }
   \uElseIf{$\mathrm{ValidatePush}(\psi) == True$}{
   $\mathrm{ClearObstacle}(\psi, push)$\;
   \textbf{Continue}
   }
   \Else{
    $\delta \gets \mathrm{FindBindingObject}(\psi)$\;
    $M_{global} \gets \mathrm{AddStaticObstacle}(M_{global}, \delta)$\;
    $R_{init} \gets R$\;
    $\mathrm{NAMOPlanner}(M_{global}, R_{init}, R_{goal})$\;
  }
  }
  \textbf{end}\\
 }
 \caption{NAMOPlanner($M_{global}, R_{init}, R_{goal}$)}
\end{algorithm}

\begin{algorithm}[tb]
\SetAlgoLined
\SetKwInOut{Input}{Input}
\SetKwInOut{Parameter}{Parameter}
\SetKwInOut{Output}{Output}
    \Input{List of primitive candidates $\Psi$, closest occupied global path coordinate $o$.}
    \Parameter{Local map inflation radius $r$.}
    \Output{Primitive that blocks the path $\psi$}
  \For{\textbf{each} $\psi \in \Psi$}{
    $\rho \gets \mathrm{GetPrimitiveCenter}(\psi)$\;
    $\mathbf{v} \gets \mathrm{Get2DVector}(\rho, o)$\;
    \uIf{$\psi.shape == Plane $}
    {
    $\mathbf{n} \gets \mathrm{GetPlaneNormal}(\psi)$\;
    $d_p \gets \mathrm{ProjectionDistance}(\mathbf{v}, \mathbf{n})$\;
    $d_r \gets \mathrm{RejectionDistance}(\mathbf{v}, \mathbf{n})$\;
    \uIf{$d_p < r \land d_r < (\psi.d/2+r)$}
    {
    \textbf{return} $\psi$;
    }
    }
    \Else
    {
    \uIf{$\|\mathbf{v}\| < (\psi.r + r)$}{
    \textbf{return} $\psi$;
    }}
    }
 \caption{FilterPrimitive($\Psi, o$)}
 
\end{algorithm}

In this study, a complete navigation framework is built integrating affordance-embedded obstacles and CITO along with standard navigation tools. The general flow is outlined in Algorithm 1. Given the global map of known static obstacles $M_{global}$, the robot's initial pose $R_{init}$, and a goal pose $R_{goal}$, the algorithm starts by calculating an optimal $path$ using $A^*$ \cite{hart1968formal}. An ego-centered local map $M_{local}$ is constructed while the robot travels along the $path$. If no obstacle is found, the robot executes the path until it reaches the goal. However, if the $path$ is blocked on $M_{local}$, the closest occupied point to the current configuration $R$ on the $path$ is recorded as $o$. The robot then detects primitives $\Psi$ in the $path$ direction using the point cloud data $P$. For all the detected primitives in view, the one that blocks the path is identified by the $FilterPrimitive$ function, which is described in Algorithm 2, and the object's \textit{movability} is validated. 

The filtering method, $FilterPrimitive$ is based on a validation vector pointing from $o$ to the primitive's center $\rho$. For a plane, the projection distance and rejection distance of the validation vector with respect to the plane normal vector is calculated. By comparing the distances with the plane's width, the robot can determine if the plane is obstructive. For cylinder and sphere, the norm of the validation vector is compared with the radius of the primitive. 

If the obstacle is validated as \textit{movable}, the robot resumes the $path$ after moving the obstacle. If the obstacle has a \textit{liftable} affordance, the path can be cleared by simply picking up the obstacle and placing it at another location. If the obstacle has only the \textit{pushable} affordance, the robot then uses CITO to find a motion that pushes the obstacle out of the way and continues tracking the $path$. Finally, if the obstacle is judged \textit{unmovable}, this new static obstacle is added to $M_{global}$, and the $path$ is re-planned.

\section{Experiments}
\subsection{Experimental Setup}
In this section, we describe the experiments carried out to demonstrate the effectiveness of our navigation framework. The experiment was performed with Toyota's Human Support Robot (HSR) \cite{yamamoto2018development, yamamoto2019development}. HSR is a holonomic-driven mobile robot equipped with a 5-DOF arm and a gripper. The sensors used during the experiments are an RGB-D camera for point cloud data collection, a laser range sensor for mapping, a wrist wrench sensor for force feedback during affordance validation. MuJoCo \cite{mujoco} is used to model the underactuated dynamics in CITO. SQOPT \cite{sqopt77} is used to solve the convex subproblems during the successive convexification process. Computations are performed on a laptop with Intel Core i7-6700HQ and Nvidia GTX 960M.

\begin{figure}[tb]
      \vspace{0.2cm}
      \centering
      \includegraphics[width=0.47\textwidth]{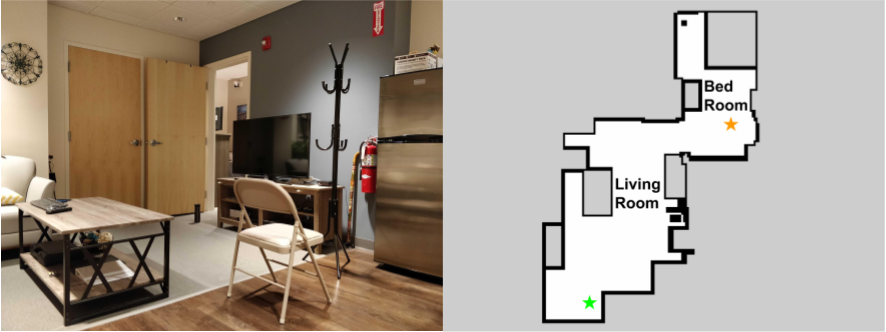}
      \caption{NUHome and the global map. The HSR is commanded to navigate from the green star to the orange star. The chair and the water bottle are not registered on the global map and the HSR is not aware of their existence beforehand.}
      \label{nuhome}
    \end{figure}

\begin{figure}[tb]
      \vspace{0.2cm}
      \centering
      \includegraphics[width=0.47\textwidth]{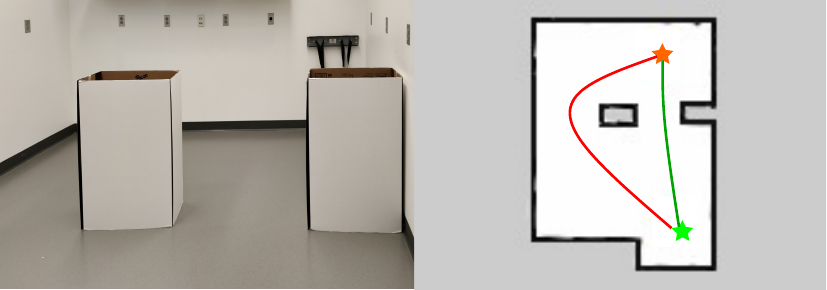}
      \caption{Rectangular room and the global map. The HSR is commanded to navigate from the green star to the orange star. The green path is shorter but blocked by a water bottle (not shown in figure). The red path is longer but clear of obstacle.}
      \label{rectangular}
    \end{figure}

Two tasks are designed for the HSR to interact with \textit{movable} and \textit{unmovable} obstacles respectively. \textbf{Task 1} is designed for the HSR to utilize affordance \textit{pushability} and \textit{liftability}. Experiments with \textbf{Task 1} are conducted in an experimental domestic environment called NUHome that is designed to simulate the actual working environment for a service robot. Fig. \ref{nuhome} shows the living room of NUHome and the global map consists of static obstacles provided to HSR. In {Task 1}, the HSR needs to navigate to the bedroom from the charging station in the living room. However, the path to the bedroom is blocked by a chair and a water bottle, and thus this task is infeasible for conventional navigation methods that only avoid obstacles. \textbf{Task 2} is designed for the HSR to perform re-planning with the presence of an \textit{unmovable} obstacle. Experiments with \textbf{Task 2} are conducted in a rectangular room with two artificial partitions. Fig.\ref{rectangular} shows the environment and the global map. The HSR needs to navigate from the green star to the orange star. Two different routes (i.e., the short green path and the long red path) are available for this task. However, the green path is blocked by a heavy water bottle while the red path is clear. The threshold value $f_L$ and $f_P$ are also lower during \textbf{Task 2} so that the water bottle is \textit{unmovable}. Unlike \textbf{Task 1}, \textbf{Task 2} is also achievable for conventional navigation methods.
The trajectories that the HSR takes to complete both \textbf{Task 1} and \textbf{Task 2} can be found in the accompanying video.

\subsection{Performance Analysis and Discussion}

\begin{figure}[t]
      \vspace{0.2cm}
      \centering
      \includegraphics[width=0.47\textwidth]{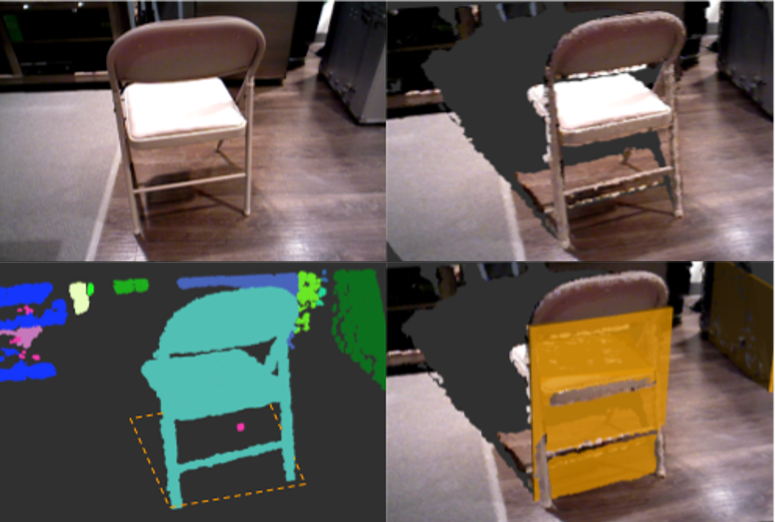}
      \caption{Affordance detection for a chair. From top left to bottom right: raw image, raw point cloud, part-base segmentation and the footprint of the chair, primitives (the orange planes) with affordance hypothesis (\textit{pushable}).}
      \label{chair}
    \end{figure}

\begin{figure}[t]
      \vspace{0.2cm}
      \centering
  \includegraphics[width=0.47\textwidth]{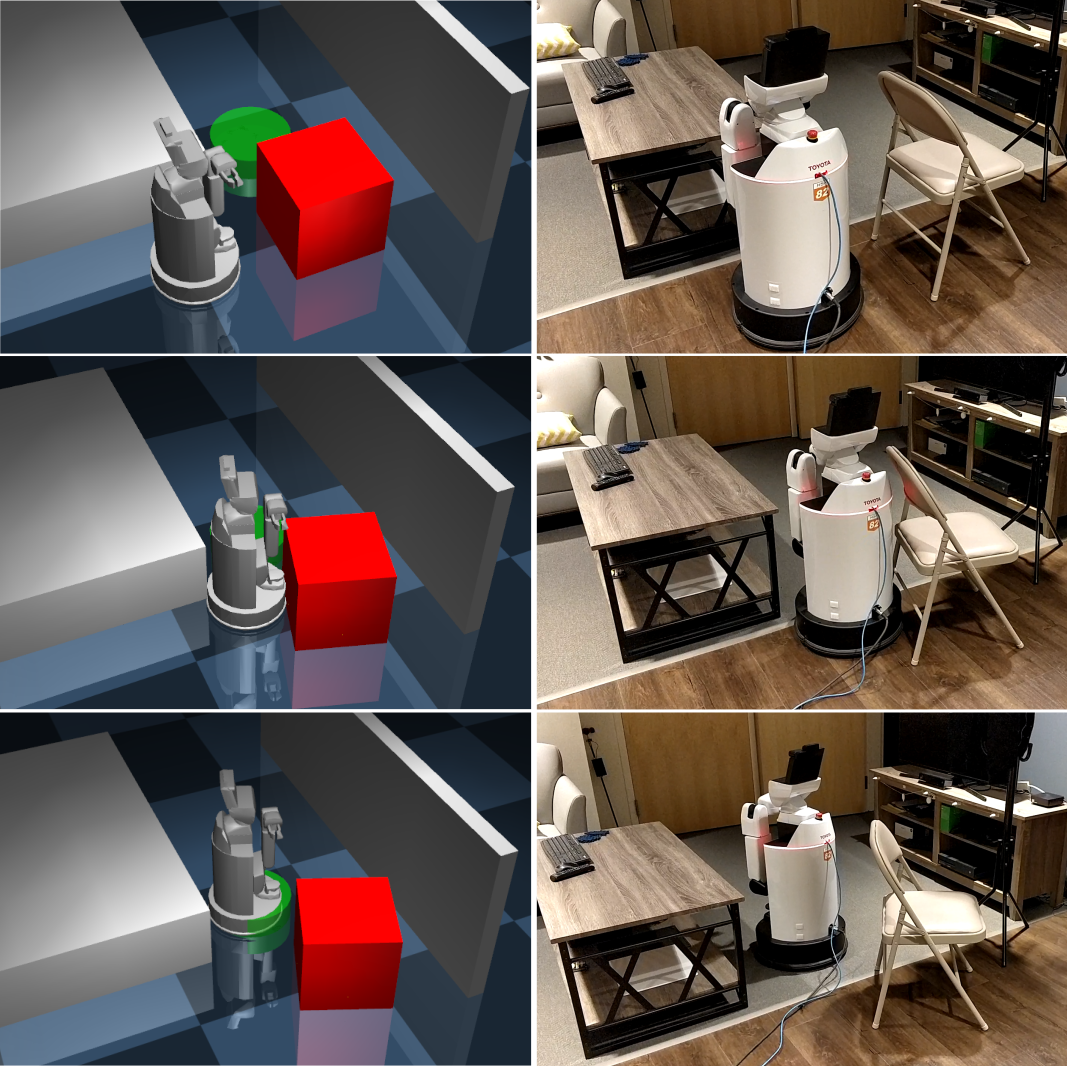}
      \caption{The HSR executes the optimized trajectory in simulation and in reality. In simulation, green spot is the goal position, red block represents the perceived movable obstacle and gray blocks represent the static obstacles that need to be avoided.}
      \label{cito}
    \end{figure}

In \textbf{Task 1}, by using the provided global map, the HSR plans a path navigating from the charging station to the bedroom. During the execution of this path, the HSR detects an unexpected chair blocking its way. An affordance detection is then performed and the chair is validated as \textit{pushable}. Fig.\ref{chair} shows the stages of visual detection. An affordance-embedded obstacle is created, which provides information about the obstacle's footprint and vertical surfaces to push. Then, a contact-interaction trajectory that pushes the chair off the path is planned by using the proposed CITO method. Fig. \ref{cito} demonstrates the HSR executing the pushing motion in simulation and experiment. Note that the orientation of the HSR is altered by 90 degrees for the hardware experiment so that the LIDAR located in the front is not blocked.

\begin{figure}[t]
      \centering
  \includegraphics[width=0.46\textwidth]{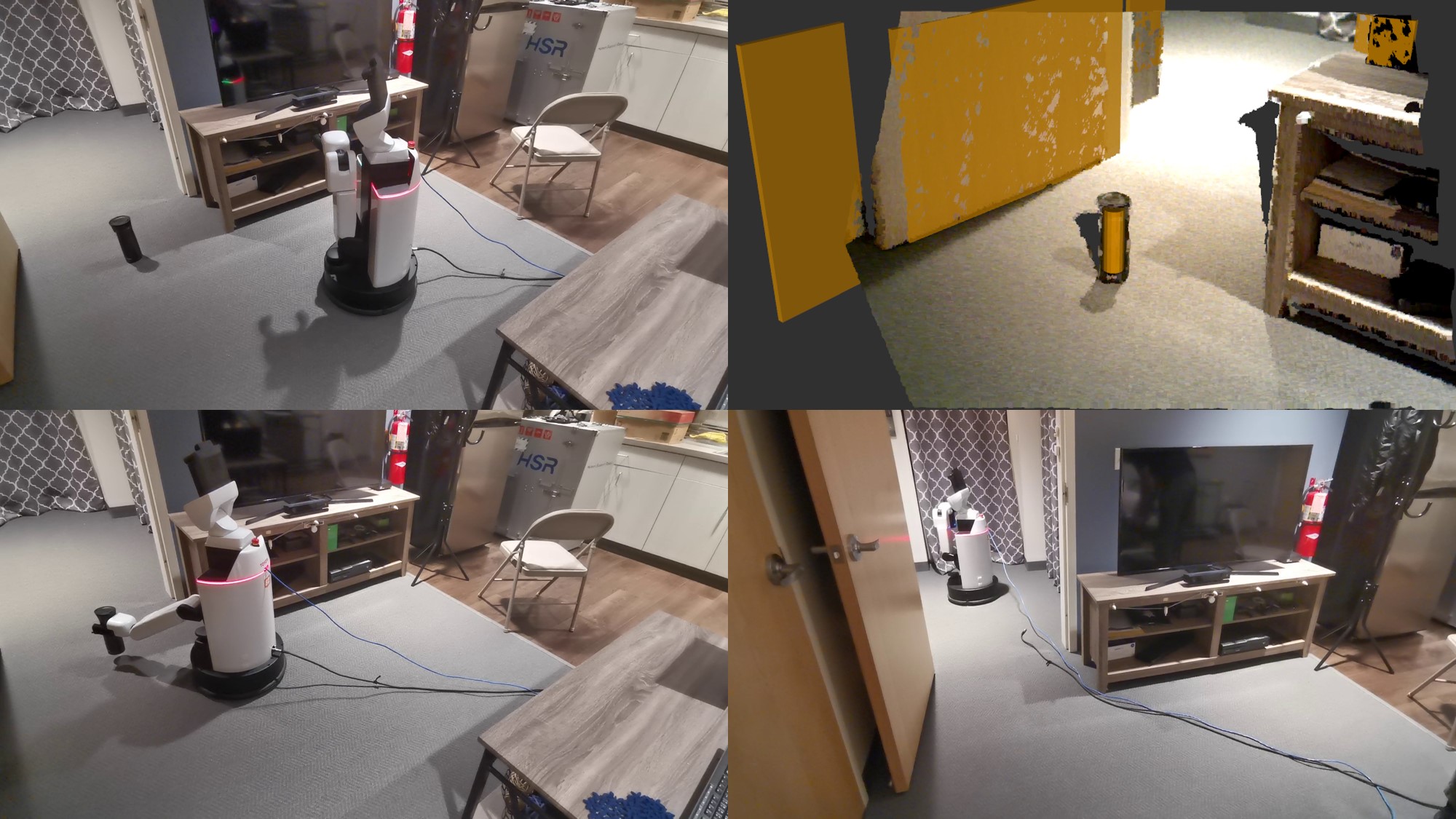}
      \caption{The HSR picks up the water bottle to clear the blocked path. From top left to bottom right: the HSR finds the blockage, affordance detection results, validating liftability, reaching the goal by clearing the path.}
      \label{bottle}
    \end{figure}

The detection and utilization of the \textit{liftability} affordance are performed when the HSR finds the water bottle that is blocking the bedroom door. The steps that the HSR takes to remove the bottle from its path are depicted in Fig.~\ref{bottle}. The results of \textbf{Task 1} prove that with the proposed navigation framework, a mobile robot can be made capable of handling unexpected obstacles on the path and completing an otherwise infeasible task, using conventional navigation methods.

In \textbf{Task 2}, the HSR first takes the green path until it detects the water bottle. The validation of \textit{liftability} and \textit{pushability} are performed consecutively and the water bottle is estimated as \textit{unmovable}.
The HSR then marks the water bottle as static obstacle on $M_{global}$ and takes the red path as a detour to the goal. The results of \textbf{Task 2} demonstrate the re-planning capability of the proposed framework with the existence of \textit{unmovable} obstacles. Admittedly, conventional navigation methods that only avoid obstacles perform better in terms of runtime for \textbf{Task 2} as the detour is taken immediately when the original path is blocked. This is a compromise of our current framework for reasoning about physical interactions with the environments. However, the payback from the reasoning enables the mobile robot to navigate in fully blocked environments like in \textbf{Task 1} or to discover a shorter path if the obstacle is instead \textit{movable} under circumstances of \textbf{Task 2}.

The computation times spent on the affordance detection, the CITO, and the execution for three experimental runs of chair pushing are presented in Table~\ref{table2}. It is shown that other than the actual execution of motions, the CITO takes most of the time. In the current setting, we run the CITO for a simulation time of 10~s and a control sampling period of 0.5~s. Although the computation time can be greatly reduced if the simulation was shorter, we choose a relatively long period to make sure the robot can always reach the goal while respecting the bounded linear and angular velocities ($\pm$2~m/s and $\pm$2~rad/s). 
   
    \begin{table}[!t]
    \centering
    \caption{Computation times for the affordance detection, the CITO, and the execution for three repetitions of navigation with chair pushing in term of seconds}
    \label{table2}
    \begin{tabular}{|c|c|c|c|c|}
    \hline
    Run & Affordance & CITO & Execution & Total \\ \hline
    1 & 59 & 69 & 69 & 197 \\ \hline
    2 & 38 & 64 & 75 & 177 \\ \hline
    3 & 35 & 62 & 85 & 182 \\ \hline
    \end{tabular}
    \vspace{0.2cm}
    \end{table}
Additionally, in the current implementation of the CITO, the derivatives are calculated by numerical differentiation, which is the bottleneck of this computation. Replacing this step by the analytical derivatives provided in \cite{carpentier2018analytical,pinocchio}, the CITO can be accelerated substantially. Another time consuming process is the affordance detection. Depending on the size of the registered point cloud, the computational time varies significantly. A better tuning of the down sampling rate would help to speed up this process.

\addtolength{\textheight}{-2cm}

\section{Conclusions}
In this paper, we present a navigation framework that incorporates an affordance embedding and contact-implicit trajectory optimization to solve the navigation among movable obstacle problem. We consider a real-world scenario where a service robot is blocked by unexpected and unknown obstacles while navigating on a given static map. We take advantage of the concept of affordances to determine the \textit{movability} of unknown obstacles based on both visual perception and physical validation. As a result, the mobile robot can assess the \textit{liftability} and \textit{pushability} of general obstacles reliably. We propose an affordance-embedding method that binds multiple affordances with an obstacles and an efficient validation procedure. Furthermore, for the \textit{pushability} affordance, we utilize a customized CITO method to plan contact-interaction trajectories to reach a desired pose without a predefined contact schedule. We demonstrate the proposed navigation framework on the HSR. 

Future work will focus on improving the navigation efficiency of the framework to handle complex environments, where multiple obstacles and detours are present. A high level task planner will be implemented to address two issues during NAMO: (\textit{i}) planning the interaction sequence with multiple obstacles and (\textit{ii}) making choice between utilizing an affordance and detouring the obstacle.





\bibliographystyle{IEEEtran}
\bibliography{citation}

\end{document}